\documentclass{article}
\usepackage{spconf,amsmath,graphicx,amssymb}
\graphicspath{{figures/}}
\usepackage{bbm}
\usepackage{xcolor}
\usepackage{booktabs}
\usepackage{tikz}
\usepackage{enumitem}

\usetikzlibrary{calc,shapes,positioning}
\usetikzlibrary{arrows}
\newcommand{\midarrow}{\tikz \draw[-triangle 90] (0,0) -- +(.1,0);}
\usepackage{mystyle}

\usepackage{bm}
\newcommand{\ubar}[1]{\mkern2mu\underline{\mkern-2mu #1\mkern-2mu}\mkern2mu}
\newcommand{\ubm}[1]{\ubar{\bm{#1}}}

\def\N{\mathbb{N}}
\def\R{\mathbb{R}}

\def\spo2{$SpO_2$}
\def\auth{3pt}

\title{Hidden Markov Models for sepsis detection in preterm infants}

\name{\parbox{\linewidth}{\centering
Antoine Honor\'e $^{\star \dagger}$,\hspace{\auth} %
Dong Liu$^{\star}$, \hspace{\auth}%
David Forsberg$^{\dagger}$, \hspace{\auth}%
Karen Coste$^{\dagger}$, \hspace{\auth}%
Eric Herlenius$^{\dagger}$\\
Saikat Chatterjee$^{\star}$, \hspace{\auth}%
Mikael Skoglund$^{\star}$\thanks{This study was supported by KTH-SLL collaborative grant HMT (2016-0764 and 20180866), the Swedish National Heart and Lung Foundation (20180505) and the Brain Foundation (FO2017-0203).}%
}
}
 \address{$^{\star}$ Div. Information Science and Engineering, KTH Royal Institute of Technology, Stockholm, Sweden\\
   $^{\dagger}$ Dept. Women's and Children's Health, Karolinska Institutet, Stockholm, Sweden}

\begin{document}
\maketitle
\begin{abstract}
We explore the use of traditional and contemporary hidden Markov models (HMMs) for sequential physiological data analysis and sepsis prediction in preterm infants. We investigate the use of classical Gaussian mixture model based HMM, and a recently proposed neural network based HMM. To improve the neural network based HMM, we propose a discriminative training approach. Experimental results show the potential of HMMs over logistic regression, support vector machine and extreme learning machine.
\end{abstract}

\begin{keywords}
Neonatal Sepsis, Hidden Markov Model
\end{keywords}

\section{Introduction}
\label{sec:intro}
Newborn babies who are under care in neonatal intensive care units (NICUs) may rapidly develop infections, including sepsis. 
It is a standard practice that the diagnosis of a sepsis is aided by the analysis of certain biomarkers from a biological sample such as blood, cerebrospinal fluid or urine. Use of biomarkers has two distinct disadvantages: abnormal levels of biomarkers occur late in the development of the disease, and biomarkers are also known to be unspecific to sepsis. This may lead to delayed treatments, long-term morbidity and death. Our interest is to explore use of physiomarkers for sepsis diagonosis. Physiomarkers are features from physiological signals, for example, respiratory signals.

Infectious diseases alter the heart and breathing patterns in infants \cite{griffinAbnormalHeartRate2003a, siljehavProstaglandinE2Mediates2015}.
These patterns include clinical events, such as Apnea-Bradycardia-Desaturation (ABD) events \cite{fairchildClinicalAssociationsImmature2016}.
These clinical events are also known to be unspecific as 97\% of extremely preterm infants suffer from ABD-events at NICUs  \cite{hofstetterCardiorespiratoryDevelopmentExtremely2008}.
In addition, the developmental age of a preterm infant affects the frequency of such events. NICU at Karolinska University hospital is equipped with bedside monitors performing a continuous recording of vital signs including respiratory frequency (RF), electrocardiogram RR-intervals (RRi) and blood oxygen saturation level (\spo2). The clinicians receive the data at $1$Hz frequency.
This is a continuously available sequential data of relevant physiological signals.

In this article we explore the use of HMMs for efficient analysis of the sequential data and prediction of sepsis in preterm infants. We directly use the raw sequential data without any manual feature extraction.
To the best of authors' knowledge this line of research has not been done before.


\noindent\textbf{Relevant literature:}
Heart Rate Observation (HeRO) system \cite{hicksHeartRateObservation2013, griffinHeartRateCharacteristics2005} is commercially available. HeRO uses features extracted from RRi signal as input to a logistic regression model for sepsis prediction. Extracted features in HeRO are mean, standard deviation and sample asymmetry of RRi signal. RALIS \cite{gurMathematicalAlgorithmDetection2014} uses age dependent thresholds on RF, RRi, \spo2, body temperature, desaturation and bradycardia events.
Pulse oximetry predictive score (POPS)  \cite{sullivanEarlyPulseOximetry2018} uses mean, standard deviation, skewness, kurtosis and min-max cross correlation between RR-interval and \spo2 to compute a risk score. POPS also uses logistic regression.

Use of sequential physiological data for sepsis prediction was recently explored in \cite{guzeyMachineLearningBased2018}. The work \cite{guzeyMachineLearningBased2018} uses feature extraction from the data and treat them as a static features for further use of machine learning. They do not explore dynamical systems such as HMM. There are two main works \cite{stanculescuAutoregressiveHiddenMarkov2014,parenteHiddenMarkovModels2018} where HMM was explored for sepsis prediction. The work \cite{stanculescuAutoregressiveHiddenMarkov2014} uses sequential clinical events such as Bradycardia-Desaturation events and does not use raw physiological signals. On the other hand, the work \cite{parenteHiddenMarkovModels2018} considers raw physiological data for adults, where HMM state distributions are modelled using kernel density estimators.  

\noindent \textbf{Our contributions:}

\begin{itemize}[noitemsep, nolistsep]
\item Motivated by end-to-end learning approaches, we explore direct use of the raw sequential physiological data without any manual feature extraction.
\item We explore classical GMM-HMM and recent Flow model based HMM \cite{liuPoweringHiddenMarkov2019}. Flow model is a neural network based distribution modeling method \cite{dinhNICENonlinearIndependent2014}.
\item For further improvement, we explore use of cross-entropy minimization based discriminative training.
\item We compare the performance of HMMs with the performance of logistic regression, support vector machine (SVM) \cite{steinwartSupportVectorMachines2008} and a popular low complexity neural network called extreme learning machine (ELM) \cite{guang-binhuangExtremeLearningMachine2012}.
\end{itemize}


\noindent \textbf{Notations:}
We define $\forall n \in \N, [n] = \{1\cdots n\}$.
$E[\cdot]$ denotes the expectation operator.
Let $\cdot^{\intercal}$ denote the transpose operation.
We denote $\forall I$ interval, $\mathbbm{1}{}_{I}$ as the indicator function of the interval.

\section{Hidden Markov Models - Traditional and Contemporary}

\label{sec:met}

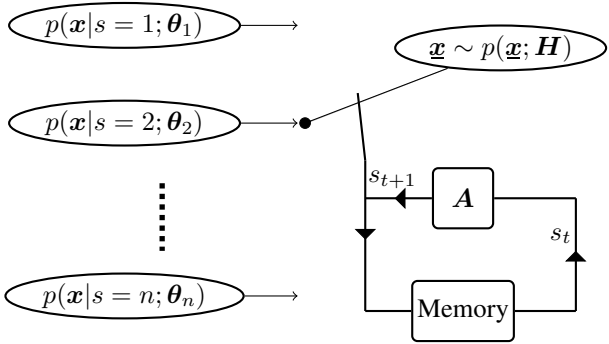
\begin{figure}[!t]
  \centering
  \begin{tikzpicture}
    \tikzstyle{enode} = [thick, draw=black, ellipse, inner sep = 1pt,  align=center]
    \tikzstyle{nnode} = [thick, rectangle, rounded corners = 2pt,minimum size = 0.8cm,draw,inner sep = 2pt]
    \node[enode] (g1) at (-0.5,1.8) {$p(\bm{x}| s=1; \bm{\theta}_{1})$};
    \node[enode] (g2) at (-0.5,0.5) {$p(\bm{x}| s=2; \bm{\theta}_{2})$};
    \node[enode] (gs) at (-0.5, -1.8) {$p(\bm{x}| s=n; \bm{\theta}_{n})$};
    \node[enode] (x) at (4.5,1.5){$\ubm{x}\sim p(\ubm{x};\bm{H})$};

    \draw[dotted,line width=2pt] (0,-0.3) -- (0,-1.2);
    \filldraw[->] (1.9, 0.5)circle (2pt) --  (x) ;
    \draw[->] (g1) -- (1.8, 1.8);
    \draw[->] (g2) -- (1.8, 0.5);
    \draw[->] (gs) -- (1.8, -1.8);

    \begin{scope}[xshift=0.5cm, thick, every node/.style={sloped,allow upside down}]
      \node[nnode] (m) at (3.5,-2) {Memory};
      \node[nnode] (a) at (3.5,-0.5) {$\bm{A}$};

      \draw (2.1,0.9)-- (2.2, 0.);
      \draw (2.2,0.)-- node {\midarrow} (2.2,-2);
      \draw (2.2,-2)-- (m);
      \draw (m)-- (5, -2);
      \draw (5, -2)-- node {\midarrow} (5 ,-0.5);
      \draw (5, -0.5) -- (a);
      \draw (a)-- node {\midarrow} (2.2, -0.5);
      \node at (4.8, -1) {$s_{t}$};
      \node at (2.56, -0.25) {$s_{t+1}$};
    \end{scope}
  \end{tikzpicture}
  \caption{HMM model illustration \cite{liuPoweringHiddenMarkov2019}}\label{fig:hmm}
  \vspace{0.3cm}
\end{figure}

HMMs have been shown to be useful in various areas of biology \cite{yoonHiddenMarkovModels2009}, and is widely used for speech recognition tasks \cite{rabinerTutorialHiddenMarkov1989,chatterjeeAuditoryModelBasedDesign2011}. A HMM $\bm{H}$, is a probabilistic model used to represent time series $\ubar{\bm{x}} = \left[ \bm{x}_1, \cdots, \bm{x}_T\right]^{\intercal}$, where $\bm{x}_t\in \RR^{N}$ is the sample at time $t$ and $T$ is the total length.
The hypothesis space of HMMs is defined as $\Hh := \{\bm{H} | \bm{H}=\{n, \bm{q}, \bm{A}, p(\bm{x}|{i}; \bm{\theta}_{i})\}\}$, where
\begin{itemize}[noitemsep, nolistsep]
\item $n\in\N$ is the number of hidden states in $\bm{H}$.
\item $\bm{q} = \left[ q_1, q_2, \cdots, q_{n}\right]^\intercal$ is the state initialization distribution, i.e. $\forall i \in [n],~q_i = p(s_{1}=i;\bm{H})$, where $s_{1}$ denotes the state at time $1$, i.e. the initial state.
\item $\bm{A} \in R_+^{n\times n}$ is the row stochastic transition matrix, i.e, $\forall i, j \in [n]\times[n]$,  $\bm{A}_{i,j} = p(s_{t+1}=j|s_{t}=i; \bm{H})$, where $s_{t}$ denotes the state at time $t$.
\item $p({\bm{x}}|{s};\bm{\theta}_{s}),~s\in [n]$, is the emission distribution of a sample $\bm{x}\in\R^N$, where $\bm{\theta}_{s}$ is the set of parameters for state $s$.
\end{itemize}

At each time instance $t$, signal $\bm{x}_t$ is assumed to be sampled from the state emission function $p(\bm{x}_t| s_t; \bm{\theta}_{s_t})$, and the sequence of $s_t$ is modeled by a Markov chain.
A HMM is depicted in Figure~\ref{fig:hmm}.

\subsection{Modeling of State Distribution}
We use two models for state distribution.
In the first case, we use Gaussian mixture model (GMM) as a traditional approach. We do not discuss further GMM based HMM as the model is well known \cite[chapter~13.2.1]{bishopPatternRecognitionMachine2006}.

In the second case we use neural network based probabilistic model called flow model \cite{dinhNICENonlinearIndependent2014}.
The flow model based HMM (Flow-HMM) has recently been explored in \cite{liuPoweringHiddenMarkov2019}.
The use of neural network based flow models allows modeling a complex data distribution while being able to compute the likelihood analytically.
The exact likelihood computation allows us to train the Flow-HMM in expectation-maximization (EM) approach.
We explain in brief the flow model architecture.

For the Flow-HMM, the state distribution $p(\bm{x}| s; \bm{\theta}_{s})$ is defined as an induced distribution by a generator $\bm{g}_{s}: \RR^{N}\rightarrow\RR^{N}$, such that $\bm{x}=\bm{g}_{s}(\bm{z})$, where $\bm{z}$ is a latent variable following a distribution with density function $q_{s}(\bm{z})$.
Generator $\bm{g}_{s}$ is parameterized by $\bm{\Phi}_{s}$.
That is, $\bm{\theta}_{s} = \{q_{s}(\bm{z}), \bm{\Phi}_{s}\}$.
Assuming $\bm{g}_{s}$ is invertible, by change of variable, we have
\begin{equation}\label{eq:changel-variable}
  p(\bm{x}| s; \bm{\theta}_{s}) = q_{s}(\bm{z})\bigg| \det\left( \pd{\bm{g}_{s}(\bm{z})}{\bm{z}} \right)\bigg|^{-1}.
\end{equation}

The parameters of the Flow-HMM are found using expectation-maximization (EM) assisted by gradient search. EM is used in the maximum likelihood sense. The details of the Flow-HMM training algorithm may be found in \cite[Algorithm 1]{liuPoweringHiddenMarkov2019}.

\subsection{Discriminative Flow-HMM (dFlow-HMM)}
Certain classes might include a large variety of patterns. 
Under such conditions, the likelihood of these models w.r.t. incoming input signals will generally be high.
To avoid classifying all the signals in these classes, we propose to re-weight using a discriminative step after the Baum-Welch iterations.
The implementation of this discriminative training step is intended as a fine tuning step after the initial Flow-HMM training.
This discriminative step is performed by maximizing the conditional probability of the correct class given an input training sample. Minimizing the cross-entropy allows us to maximize the quantity:
\begin{equation} \label{eq:discriminative}
\sum_{(\ubm{x},\bm{y})} \log \frac{p(\ubm{x}\mid\bm{H}_{\bm{y}})p(\bm{H}_{\bm{y}})}{\sum_{j=0}^1p(\ubm{x}\mid\bm{H}_{j})p(\bm{H}_{j})}
\end{equation}
where, $\ubm{x}= \left[ \bm{x}_1, \cdots, \bm{x}_T\right]^{\intercal}$ with $\bm{x}_t\in \RR^{N},~t=1\dots T$ is an input training sequence and $\bm{y}\in\{0,1\}$ is its class label.
The classes prior probabilities $p(\bm{H}_{i}),~i=1,2$ are infered from the training dataset.
The probability $p(\ubm{x}\mid\bm{H}_{i}),~i=1,2$ of an observed sequence $\ubm{x}$ is computed using the forward algorithm.

Computing the denominator of (\ref{eq:discriminative}) exactly, requires that all training sequences are evaluated by each class HMM.
In practice, efficiently computing the denominator is infeasible when the number of classes is too large.
In our case, we only have two classes and the computation overhead is manageable.
In our implementation, we only update the Flow-models parameters during the maximization of Equation~(\ref{eq:discriminative}).
The Markov chain parameters are left unchanged in the discriminative training phase.

We perform the optimization using a stochastic gradient descent approach.
The stochastic gradient descent algorithm used here is ADAM \cite{kingmaAdamMethodStochastic2014}.

In addition to discriminative training, 
we use feature extension technique to improve the performance of HMMs.
Feature extension was shown to improve the performances of both GMM-HMM and Flow-HMM in practices \cite{liuPoweringHiddenMarkov2019}. Our feature extension consists in concatenating raw inputs with their $1^{st}$ and $2^{nd}$ order derivatives.
The details about performance comparison with and without feature extension would be shown in the following section.


\section{Experimental Results}
\label{sec:res}
\subsection{Patient Dataset}
The bedside monitor signals of $48$ premature infants which have been under care at a NICU have been collected.
The signals used are the Respiratory Frequency (RF), the beat to beat interval (RRi) and the blood oxygen saturation level (\spo2).
All signals were sampled at $1$Hz and segmented into $20$ minutes time frames.
Time frames containing missing data were discarded.
Each time frame was then labelled based on information retrieved from the Electronic Health Records (EHR).
Similarly to HeRO and RALIS, we aim at detecting septic events earlier than clinical suspicion of sepsis, defined as the sampling of a blood culture.
In our study we use a threshold of $72$h prior to blood sample, to label a time frame as "septic".
This is in accordance with practices and results from \cite{gurMathematicalAlgorithmDetection2014} where the RALIS system, was able to trigger sepsis alarms 2.5 days earlier in a subgroup of patients.
A time frame was retro-actively labeled $1$ if it occurred at most $72$h prior to clinical suspicion.
A time frame was labeled $0$ if it occured during a day when no notes were entered in the infant's EHR.
All time frames not labeled either $0$ or $1$ were discarded.

Our final dataset consists of $22$ patients, among which $13$ males and $9$ females.
The birth weight was $1.61\pm1.10$ kg and the gestational age at birth $30.9\pm6.14$.
Our dataset consisted in $3501$ time frames, among which 1774 with label $0$ and $1727$ with label $1$.
All time frames have a constant size of $T=1200$ samples and are $3$-dimensional.

\subsection{Baseline Methods}
Here we use a baseline model similar to the HeRO model.
This model consists in a feature extraction block, followed by a logistic regression.
First, the Heart Rate Characteristic index (HRCi) is computed on all time frames.
The HRCi computation is summarized by the map 
\begin{align*}
h:\ubm{x}	 \mapsto (E[\ubm{x}_{RRi}],E[(\ubm{x}_{RRi}-E[\ubm{x}_{RRi}])^2],s(\ubm{x}_{RRi})) \in \R^3,
\end{align*}
where $\ubm{x}_{RRi}\in\R^T$ denotes the RRi signal of a time frame $\ubm{x}$.
$s$ is the sample asymmetry, defined as:
\begin{align*}
s: x   \mapsto  \frac{\sum_{i=1}^T \xi_i(x)^2\mathbbm{1}{}_{\R_+}(\xi_i(x))}{\sum_{i=1}^T\xi_i(x)^2\mathbbm{1}{}_{\R_{*-}}(\xi_i(x))}\in \R,
\end{align*}
where $\forall x\in\R^T~\xi_i(x) = x_i - \text{median}(x)$.
Note that the HeRO system only requires the RRi signal.
Here, we do not require sparsity of the results, therefore, we trained a logistic regression model with $l_2$-regularization.
The regularization parameter was learned using a 3-fold cross-validation grid search in the interval $\{10^{-5},\cdots,10^{5}\}$.
Additionally we used a second set of features known as POPS \cite{sullivanEarlyPulseOximetry2018} which uses: the mean, the standard deviation, the skewness and the kurtosis of the RRi and \spo2, together with the min-max cross correlation between the RRi and \spo2 with a limited lag of 30 seconds.
The results associated with these two sets of features are presented in Table~\ref{table:results:Lin}.

\subsection{Experiments}
We performed binary classification of two types of fixed length input time series in a maximum likelihood framework with GMM, Flow and dFlow -HMMs as probabilistic models.
As a baseline, we used the clinically used HRCi index and the more recent POPS features as input to a logistic regression system.
We repeated our experiments $3$ times and each time a random $30\%$ of the patients was left out for testing.
This lead to $2361 \pm 353$ time series in the training sets and $1140 \pm 353$ time series in the testing sets.
The code was written in Python using the Scikit-learn library for the HeRO system, hmmlearn for GMM-HMM,  and PyTorch to implement Flow and dFlow-HMM.
The GMM-HMM hyper-parameters were the number of states and the number of Gaussians per state.
The number of Gaussians per states was varied between \{2,4,6,8,10,12\}.
For Flow and dFlow-HMMs the hyper-parameters were the number of states, the number of Flow-model per state, the number of chains in each Flow-model, the size of the networks in the coupling layers.
Given our limited input dimension, the size of the networks in the coupling layers was fixed to $3$.
The number of chains in the coupling layer of each Flow-model was varied between $4$ and $8$.
We varied the number of states in our HMMs in $\{3,6,9\}$ and show the results in Table~\ref{table:results:HMM}.
For the logistic regression, optimal regularization parameter was found with cross-validation and grid search in the set $\{10^{-5}, \cdots, 10^{5}\}$.
We used a different set of input features to test our models in different conditions.
HMMs are trained on raw time series and on raw time series with first and second order derivatives.
The logistic regression model is trained on HRCi, $3$-dimensional feature, and on POPS, $10$-dimensional features.

\subsection{Numerical Results}
The results for the linear prediction systems are presented in Table~\ref{table:results:Lin} and the results for the HMMs are presented in Table~\ref{table:results:HMM}.
The bold fonts corresponds to the maximum performance across HMM models given a number of states.

\begin{table}[!ht]
\center
\caption{Test accuracy of HMMs}
\setlength{\tabcolsep}{4pt}
\begin{tabular}{c|c|c|c}    \toprule
Number of states 		&  n=3 			    				&  n=6									&  n=9 						\\  \bottomrule
                                        \multicolumn{4}{c}{Raw time series} 		\\
\hline
GMM-HMM				&   0.68 $\pm$ 0.03 		&    \textbf{0.68} $\pm$ 0.03 			&   	\textbf{0.69} $\pm$ 0.03  	\\
FlowHMM  				&   0.67 $\pm$ 0.04     	&    0.61 $\pm$ 0.08   		    &   	0.63 $\pm$ 0.08   \\
dFlowHMM 				&   \textbf{0.70} $\pm$ 0.10    	&	   0.67 $\pm$ 0.06      	    &  	    0.65 $\pm$ 0.04    \\
   \toprule
                                        \multicolumn{4}{c}{Raw time series + $1^{st}$ and $2^{nd}$ order derivatives}\\
\hline
GMM-HMM				&   \textbf{0.75} $\pm$ 0.05 		&    \textbf{0.74} $\pm$ 0.08			&   	\textbf{0.74} $\pm$ 0.05  	\\
FlowHMM  				&   0.69 $\pm$ 0.07    	&    0.66 $\pm$ 0.06  		    &   	0.59 $\pm$ 0.08  \\
dFlowHMM 				&  0.71 $\pm$ 0.04    		&	   0.72 $\pm$ 0.10     	    &  	   0.67 $\pm$ 0.04   \\
\hline
\end{tabular}

\label{table:results:HMM}
\end{table}

\begin{table}[!ht]
\center
\caption{Comparison of HMM with other models}
\begin{tabular}{*4c}
\hline
 Model							&   			    & Variety 		 	\\  \bottomrule
\hline
\multicolumn{4}{c}{Using features}    \\ \hline
Logistic Regression	 & & HeRO			&   0.52$\pm$0.08 \\
& & POPS & 0.57$\pm$0.04				 \\   \hline
\multicolumn{3}{c}{Using raw sequential data}    \\    \hline  
SVM      & &                       &   0.60 $\pm$ 0.04   			\\
ELM		& &						&   0.60 $\pm$ 0.03   				\\ 
dFlowHMM     & &                     &   0.70 $\pm$ 0.10  \\ \hline

\hline
\end{tabular}
\label{table:results:Lin}
\end{table}

The HeRO (HRCi + logistic regression) does not perform as well as expected with only $52\%$ of correct classifications. 
With POPS features, the performance of the linear classifier reaches $57\%$.
As expected, the linear classifiers are outperformed by Gaussian kernel SVM, ELM and our HMMs.
SVM and ELM both reach an accuracy of $60\%$ and comparable standard deviations of $4\%$ and $3\%$.
This is lower than dFlow-HMM which reaches $70\%$ accuracy and outperforms both Flow-HMM $67\%$, and GMM-HMM $68\%$ when the number of states $n=3$ and for raw-time series.
These results are contrasted by the large standard deviation of dFlow-HMM $10\%$, which is larger than both Flow-HMM $4\%$ and GMM-HMM $3\%$.
When the number of states increases to $n=6$ and $n=9$, GMM-HMM reaches $68\%$ and $69\%$ which outperforms Flow-HMM with $61\%$ and $63\%$, and dFlow-HMM with $67\%$ and $65\%$.
When the input time series is augmented with $1^{st}$ and $2^{nd}$ order derivatives, GMM-HMM reaches its highest performance with $75\%$ accuracy at $n=3$.
Flow-HMM and dFlow-HMM also reach their highest performance with $69\%$ at $n=3$ and $72\%$ at $n=6$.
Here, GMM-HMM outperforms both Flow and dFlow-HMM.

\subsection{Discussion}
The HeRO model surprisingly under-performs on our dataset.
This poor performance may be due to inconsistencies in the sepsis definition.
The definition of a sepsis varies between the different studies, here we include culture positive and culture negative which differs from the initial HERO study \cite{griffinHeartRateCharacteristics2005} but is in accordance with the definitions used in the more recent RALIS study \cite{mithalComputerizedVitalSigns2016}.

The performance of GMM-HMM is significantly increased when adding $1^{st}$ and $2^{nd}$ order derivative as part of the input.
This is in accordance with the initial studies performed on speech processing  tasks \cite{liuPoweringHiddenMarkov2019}.
Our attempt to improve the performance of Flow-HMM using discriminative training was successful for both raw time series and $1^{st}$ and $2^{nd}$ order derivatives inputs.
This is encouraging, given that our current discriminative training consists of only one epoch.
We expect the marginal gain of discriminative training to improve the performance of dFlow-HMM even further with more iterations.
However, we note a decrease in performance of Flow-HMM and dFlow-HMM as the number of hidden states is increased. This is due to the fact that there are more parameters to learn as we increase the hidden states, for the same amount of data.
Among all the tested systems, GMM-HMM has the smallest standard deviation, which indicates that this model is robust to changes in training dataset.
Flow models were originally designed for high-dimensional data distribution modeling.
As expected, additional dimensions in Flow-HMM input lead to improvement in performances compared to the $3$-dimensional case.
We conjecture that Flow-HMM and dFlow-HMM suffer from insufficient training data, considering the fact that more parameters are to be learnt than GMM.


\section{Conclusion}
We studied the performance of Hidden Markov Models compared to state-of-art logistic regression based classification models for neonatal sepsis detection.
We showed that, on our dataset consisting of $22$ patients, neonatal sepsis detection may be enhanced with the use of Hidden Markov Models.
We observed that Gaussian Mixture Models for state emission probability distributions performs well, and with a low standard deviation compared to other models.
dFlow-HMM was shown to outperform GMM-HMM in a limited number of scenarios.
This may constitute an important building block in the future design of Flow-model based Hidden Markov Model.
Adding the derivatives of the signal as an input, lead to improvement of the HMMs.
Our study paves the way for further research on Hidden Markov Models topology which may lead to improved neonatal sepsis detection in NICUs.

\section{Acknowledgements}
This project was approved by the regional ethics review board (2011/1891-31/2).
Informed consent was provided by all parents/guardians.
\vfill\pagebreak
\bibliographystyle{IEEEbib}
\bibliography{strings,refs,ICASSP2020}
\end{document}